\documentclass{article}
\usepackage[preprint]{log_2023}			%

\usepackage{booktabs}           %
\usepackage{multirow}           %
\usepackage{amsfonts}           %
\usepackage{graphicx}           %
\usepackage{duckuments}         %

\usepackage{soul}
\usepackage{url}
\usepackage[utf8]{inputenc}
\usepackage{caption}
\usepackage{amsmath}
\newcommand\numberthis{\addtocounter{equation}{1}\tag{\theequation}}
\usepackage{booktabs}
\usepackage{placeins}
\usepackage{float}
\usepackage{enumitem}
\usepackage{url}
\usepackage[english]{babel}
\usepackage{blindtext}

\usepackage{soul}
\usepackage{mathtools}
\usepackage{amsmath}
\usepackage{amssymb}
\usepackage{todonotes} 
\usepackage{xcolor}
\usepackage{color}
\usepackage{tabularx}
\usepackage{makecell}
\usepackage{caption}
\usepackage{subcaption}
\usepackage{algorithm}
\usepackage[noend]{algpseudocode}

\usepackage{siunitx}
\usepackage{amsmath,amsfonts,bm}

\def\eqref#1{equation~\ref{#1}}

\def\1{\bm{1}}

\DeclareMathAlphabet{\mathsfit}{\encodingdefault}{\sfdefault}{m}{sl}
\SetMathAlphabet{\mathsfit}{bold}{\encodingdefault}{\sfdefault}{bx}{n}

\newcommand{\R}{\mathbb{R}}

\newcommand{\Real}{\ensuremath{\mathbb{R}}}

\usepackage[numbers,compress,sort]{natbib}

\title{KGEx: Explaining Knowledge Graph Embeddings via Subgraph Sampling and Knowledge Distillation}

\author[Vasileios Baltatzis, Luca Costabello]{%
Vasileios Baltatzis\\
King's College, London\\
\email{vasileios.baltatzis@kcl.ac.uk}\And
Luca Costabello\\
Accenture Labs, Dublin\\
\email{luca.costabello@accenture.com}
}

\begin{document}

\maketitle         

\begin{abstract}

Despite being the go-to choice for link prediction on knowledge graphs, research on interpretability of knowledge graph embeddings (KGE) has been relatively unexplored.
We present KGEx, a novel post-hoc method that explains individual link predictions by drawing inspiration from surrogate models research. Given a target triple to predict, KGEx trains surrogate KGE models that we use to identify important training triples.
To gauge the impact of a training triple, we sample random portions of the target triple neighborhood and we train multiple surrogate KGE models on each of them. To ensure faithfulness, each surrogate is trained by distilling knowledge from the original KGE model.
We then assess how well surrogates predict the target triple being explained, the intuition being that those leading to faithful predictions have been trained on ``impactful'' neighborhood samples. Under this assumption, we then harvest triples that appear frequently across impactful neighborhoods. 
We conduct extensive experiments on two publicly available datasets, to demonstrate that KGEx is capable of providing explanations faithful to the black-box model.
\end{abstract}

\blindmathtrue

\section{Introduction}\label{sec:intro}

Knowledge graphs are knowledge bases whose facts are labeled, directed edges between entities. 
Research led to broad-scope graphs such as DBpedia~\cite{auer2007dbpedia}, WordNet, and YAGO~\cite{suchanek2007yago}. Countless domain-specific knowledge graphs have also been published on the web, from bioinformatics to retail \cite{hogan2021knowledge}.

Knowledge graph embeddings (KGE) are a family of graph representation learning methods that learn vector representations of nodes and edges of a knowledge graph. They are widely used in graph completion, knowledge discovery, entity resolution, and link-based clustering~\cite{nickel2016review}. 

\begin{figure}
\centering
\includegraphics[scale=0.125]{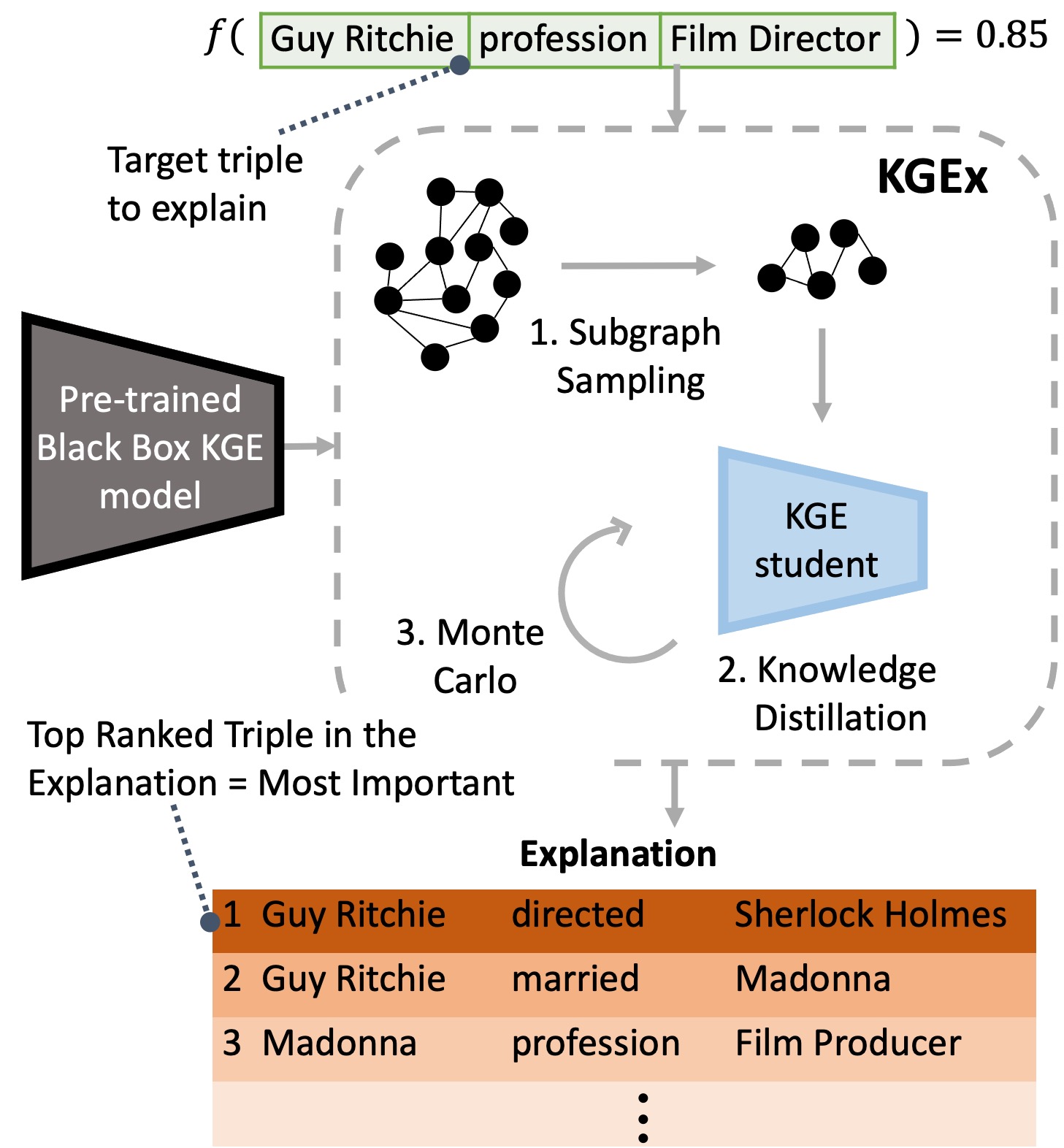}
\caption{Overview of the proposed framework. Given a pre-trained KGE model and a target triple, KGEx outputs an explanation for the prediction in the format of a list of ranked triples.}
\label{fig:hero}
\end{figure}

Despite achieving excellent trade-off between predictive power and scalability, these neural architectures suffer from poor human interpretability, to the detriment of user trust, troubleshooting, and compliance.

Previous work in knowledge graph representation learning aims at designing natively interpretable KGE models or generating post-hoc explanations for existing knowledge graph embedding models. 
Nevertheless, the field is still in its infancy and recently proposed explanation methods do not scale beyond toy datasets or do not provide thorough empirical evidence of being faithful to the KGE model being explained.

In this work, we propose KGEx, a post-hoc, local explanation sub-system for KGE models (Figure~\ref{fig:hero}). 
KGEx works with any existing KGE model proposed in literature: given a target triple predicted with a KGE model, we return an explanation in the form of a ranked list of relevant triples from the training set. We use a combination of subgraph sampling and knowledge distillation that we refine with Monte Carlo sampling. Our experiments show that KGEx provides faithful explanations that can be used beyond toy knowledge graphs.
\section{Related Work}\label{sec:relatedwork}

\textbf{Knowledge Graph Embeddings.}
Knowledge graph embedding models (KGE) are neural architectures designed to predict missing links between entities.
TransE~\cite{bordes2013translating} is the forerunner of distance-based KGE models, and inspired a number of models commonly referred to as TransX.
The symmetric bilinear-diagonal model DistMult~\cite{yang2014embedding} paved the way for its asymmetric evolutions in the complex space, ComplEx~\cite{trouillon2016complex} and RotatE~\cite{sun2019rotate}.
Some models such as RESCAL~\cite{nickel2011three}, TuckER~\cite{balavzevic2019tucker}, and SimplE~\cite{kazemi_simplE} rely on different tensor decomposition techniques. 
Models such as ConvE~\cite{DBLP:conf/aaai/DettmersMS018} or ConvKB~\cite{nguyen2018novel} leverage convolutional layers. 
Attention is used by~\cite{nathani2019learning}. The recent NodePiece uses an anchor-based approach to map entities and relations to a fixed-sized, memory-efficient vocabulary~\cite{galkin2021nodepiece}.
Recent surveys provide a good coverage of the landscape~\cite{bianchi2020knowledge}.

\textbf{Explanations for KGEs.}
Recent studies address the problem of making KGE architectures more interpretable. MINERVA \cite{das2018go}, NeuralLP \cite{yang2017differentiable}, CTPs \cite{minervini2020learning} integrate rule-based systems to deliver natively interpretable link prediction methods. Although promising, these works do not scale beyond toy datasets.
Other works consist instead of \textit{post-hoc} approaches (they operate on black-box, pre-trained KGE models) and are \textit{local} methods, i.e. they explain the prediction of a single instance (i.e. a single missing link between two entities).
Gradient Rollback~\cite{lawrence2021explaining} returns a ranked list of influential triples for a target prediction. Such list is computed by storing gradient updates during training, to the detriment of memory footprint and training time overhead.
OXKBC creates post-hoc explanations by leveraging entity and path similarities that are selected as template by an end-to-end method~\cite{nandwani2020oxkbc}.
Earlier work identifies training triples that, when removed, decrease the predicted probability score. ExplainE is grounded on counterfactual explanations and operates on toy knowledge graphs and low-dimensional embeddings \cite{kang2019explaine}. 
~\cite{zhang2019interaction} instead randomly perturbs the neighborhood of the target triple, but its design rationale is geared towards adversarial attacks rather than explainability.
\section{Preliminaries}\label{sec:background} 

\paragraph{Knowledge Graph.}
A knowledge graph $\mathcal{G}=\{ (s,p,o)\} \subseteq \mathcal{E} \times \mathcal{R} \times  \mathcal{E}$ is a set of triples $t=(s,p,o)$ each including a subject $s \in \mathcal{E}$, a predicate $p \in \mathcal{R}$, and an object $o \in \mathcal{E}$. $\mathcal{E}$ and $\mathcal{R}$ are the sets of all entities and relation types of $\mathcal{G}$.

\paragraph{Knowledge Graph Embedding Models.}
KGE encode both entities $\mathcal{E}$ and relations $\mathcal{R}$ into low-dimensional, continuous vectors $\in \R^k$ (i.e, the embeddings). 
Embeddings are learned by training a neural architecture over a training knowledge graph $\mathcal{G}$: an input layer feeds training triples, and %
a scoring layer $f(t)$ assigns plausibility scores to each triple. $f(t)$ is designed to assign high scores to positive triples and low scores to negative \textit{corruptions}. 
Corruptions are synthetic negative triples generated by a corruption generation layer:
we define a corruption of $t$ as $t^-=(s,p,o')$ or $t^-=(s',p,o)$ where $s', o'$ are respectively subject or object corruptions, i.e. other entities randomly selected from $\mathcal{E}$ \cite{bordes2013translating}. %
Finally, a loss layer $\mathcal{L}_{KGE}$ optimizes the embeddings by learning optimal embeddings, such that at inference time the scoring function $f(t)$ assigns high scores to triples likely to be correct and low scores to triples unlikely to be true.

\textbf{Link Prediction.} 
The task of predicting unseen triples in knowledge graphs is formalized in literature as a learning to rank problem, where the objective is learning a scoring function $f(t=(s, p, o)): \mathcal{E} \times \mathcal{R} \times \mathcal{E} \rightarrow \Real$ that given an input triple $t=(s,p,o)$ assigns a score $f(t) \in \Real$ proportional to the likelihood that the fact $t$ is true. 
Such predictions are ranked against predictions from synthetic corruptions, to gauge how well the model tells positives from negatives.

\textbf{Knowledge Distillation (KD).}
This method has been introduced to alleviate computational costs and allow knowledge to be transferred from large, complex models (teacher) to smaller, compact ones (student) \cite{hinton2015distilling}. Let $\mathcal{X}$ be the input data distribution, with $x_i \sim \mathcal{X}$ distinct samples drawn from that distribution. For brevity, we denote teacher and student representations as $\mathbf{g}_{T,i} = g_T(x_i) $ and $\mathbf{g}_{S,i}=g_S(x_i)$, respectively.
Conventional KD can take up the the form of Eq. \ref{eq:KD}:
\begin{equation}
\mathcal{L}_{KD} = \sum_{x_i \sim \mathcal{X}} l (\mathbf{g}_{T,i}, \mathbf{g}_{S,i}),
\label{eq:KD}
\end{equation}
where $l$ is a loss function such as the Kullback-Leibler divergence, which tries to match the representations of teacher and student for an individual sample $x_i$.

Relational KD's (RKD) \cite{Park2019RelationalKD} purpose is to transfer the relationship between individual samples from the teacher to the student, as described in Eq. \ref{eq:RKD}:

\begin{equation}
\mathcal{L}_{RKD} = \smashoperator{\sum_{(x_i,...,x_n) \sim \mathcal{X}}} l_\delta (\phi(\mathbf{g}_{T,i},...,\mathbf{g}_{T,n}), \phi(\mathbf{g}_{S,i},...,\mathbf{g}_{S,n})),
\label{eq:RKD}
\end{equation}
where $\phi$ is a relational potential function and $l_\delta$ is the Huber loss, which is defined in Eq. \ref{eq:huber}:

\begin{equation}
l_\delta (a, b) =  \\
\begin{cases}
    \frac{1}{2} (a-b)^2 &\text{for }\mid a-b \mid \leq 1,  \\
     \mid a-b \mid - \frac{1}{2}, &\text{otherwise}. \\
\end{cases}
\label{eq:huber}
\end{equation}

A particular case of relational potential function is the angle-wise relational potential $\phi_A$, which can be applied on a triple of samples and is defined as in Eq. \ref{eq:phi}:

\begin{equation}
\phi_{A}(\mathbf{g}_{T,i}, \mathbf{g}_{T,j}, \mathbf{g}_{T,k}) = \langle\mathbf{d}_{ij},\mathbf{d}_{jk}\rangle,
\label{eq:phi}
\end{equation}
where $ \mathbf{d}_{ij}=\frac{\mathbf{g}_{T,i} - \mathbf{g}_{T,j}}{\parallel\mathbf{g}_{T,i} - \mathbf{g}_{T,j}\parallel_2}$, $\mathbf{d}_{jk}=\frac{\mathbf{g}_{T,j} - \mathbf{g}_{T,k}}{\parallel\mathbf{g}_{T,j} - \mathbf{g}_{T,k}\parallel_2}$ 
and $\langle\cdot{,}\cdot\rangle$ is the dot product.

\section{Methods}\label{sec:method}
Given a pre-trained black-box KGE model and the prediction for an unseen target triple, we generate an explanation for such prediction in the form of a list of training triples ranked by their `influence' on the prediction.
Such explanation is generated by KGEx, our proposed explanation subsystem, which consists of three components. The first step samples a subgraph $\mathcal{H}$ from the original knowledge graph $\mathcal{G}$ in order to limit the search space for possible explanations (section \ref{method:subgraph}). To increase the faithfulness of the surrogate model, in the second step we utilize KD to train a new KGE model on the subgraph, while the black-box KGE model we are explaining plays the role of the teacher (section \ref{method:kd}). Finally, we repeat the second step through a Monte Carlo (MC) process. This is done to rank the triples in the subgraph according to their contribution to the prediction (section \ref{method:mc}).
\subsection{Subgraph sampling} \label{method:subgraph}
\begin{figure*}
\centering
\includegraphics[width=.9\textwidth]{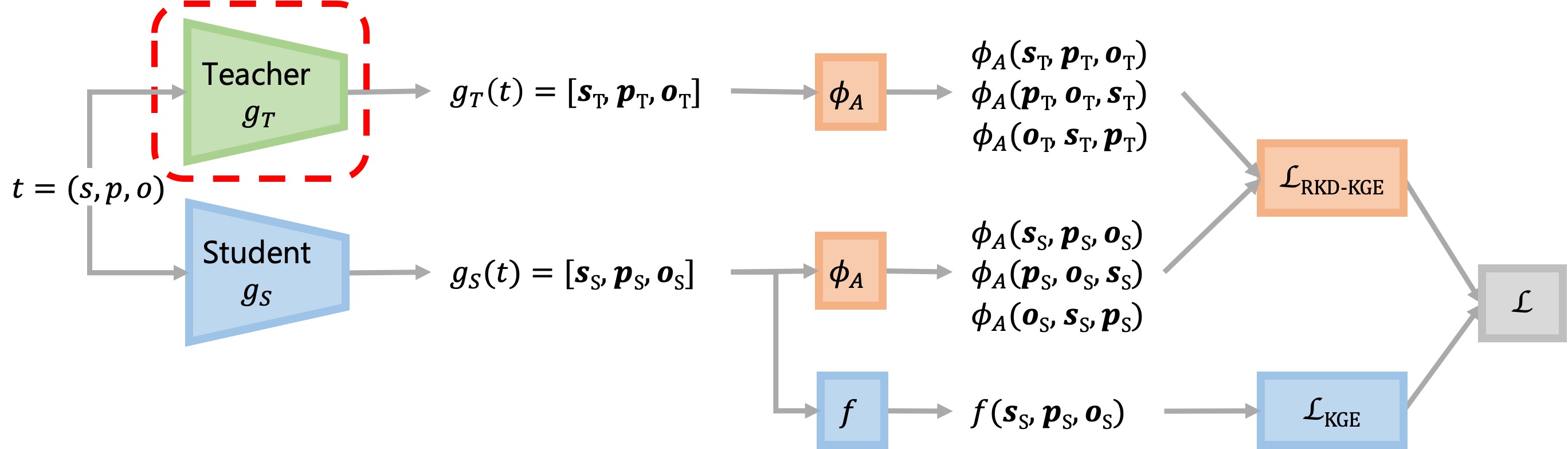}
\caption{A KGE model architecture incorporating the adapted Relational Knowledge Distillation (RKD-KGE).}
\label{fig:arch}
\end{figure*}

Our motivation stems from the explainable AI subfield of surrogate models \cite{alaa2019demystifying}. Concretely, the idea behind surrogate models is to convert a “black-box” model $g_T$ into a more interpretable “white-model” $g_S$. The main challenge when trying to design a surrogate model for a KGE model is that KGEs are transductive models and therefore have no inference capabilities, i.e. given a triple $t$ that contains an - unseen during training - entity, there is no function $g_T$, so that we can infer an output $y=g_T(t)$. Given this limitation we cannot replace $g_T$ with an interpretable $g_S$, directly. However, we can find a subgraph that will allow us to train such a surrogate model.

We formulate this task as finding the smallest subgraph $\mathcal{H} \subset \mathcal{G}$, which if used to train a KGE model $g_S$ will give a latent space representation to the target triple that is as close as possible to the one assigned by the black-box KGE model $g_T$, which was trained on the whole knowledge graph $\mathcal{G}$. While searching for $\mathcal{H}$, we are facing a trade-off that has to do with the subgraph size: a larger size promotes fidelity (i.e. faithfulness to the original model), while a smaller size reduces cognitive load and therefore favors interpretability.  

The search for the subgraph $\mathcal{H}$ of particular target triple $t^*=(s^*,p^*,o^*)$ can be broken down to two parts. The first part involves retaining the 1-hop neighborhood $N_\mathcal{G}(s^*,o^*)$ of the subject $s^*$ and the object $o^*$ of $t^*$. Using the case depicted in Fig. \ref{fig:hero}, $N_\mathcal{G}$(\textit{Guy Ritchie, Film  Director}) would include all triples involving either \textit{Guy Ritchie} or \textit{Film Director}. These are the triples that are in the vicinity of the entities of the target triple and therefore will likely play an important role in the representation that the KGE will learn for these entities. As such, the fact that \textit{(Guy Ritchie, director, Sherlock Holmes)} is important in explaining \textit{(Guy Ritchie, profession, Film Director)}. If we were to retain only this 1-hop neighborhood, we would not be able to incorporate information on any long-range (in terms of hops in the graph) information that might be pivotal for the latent representation of the target triple. Additionally, the 1-hop neighborhood of the subject and the object does not take explicitly into account the predicate $p^*$ of $t^*$, which could lead to not learning a meaningful representation for $p^*$. In the running example, the fact  \textit{(Madonna, profession, Film Producer)} might seem irrelevant at first sight and would not be part of $N_\mathcal{G}$(\textit{Guy Ritchie, Film  Director}). However, combined with the fact that \textit{(Guy Ritchie, married, Madonna)} could again lead to the target triple being predicted as positive. In the second part, to alleviate the issues above, we propose two alternatives:

\begin{itemize}
  \item \textbf{Random Walk (RW) Sampling} Apart from the 1-hop neighborhood, we also add a naive random walk of predefined size measured in numbers of steps, which starts from the target triple (see Algorithm \ref{alg:random_walk_subgraph} in Appendix~\ref{app:random_walk_algorithm}).
  \item \textbf{Predicate Neighborhood (PN) Sampling} To ensure that the predicate $p^*$ of $t^*$ is part of the subgraph we randomly sample from $\mathcal{G}$ a predefined number $n$ of triples  $t^*=(\hat{s},p^*,\hat{o})$, which have the same predicate $p^*$, and include these along with their own 1-hop neighborhoods $N_\mathcal{G}(\hat{s},\hat{o})$ in the subgraph (Algorithm~\ref{alg:predicate_subgraph}). In our example, that would entail sampling triples that involved the predicate \textit{profession}. %
\end{itemize}

\begin{algorithm}
\caption{Subgraph sampling w/ Predicate Neighborhood}\label{alg:predicate_subgraph}
\begin{algorithmic}[1]
\State \textbf{Input:} {target triple $(s^*,p^*,o^*)$, number of predicate neighbors $n$}
\State \textbf{Output:} {Subgraph $\mathcal{H}$}
\State $\mathcal{H} \gets \emptyset$
\State $N_\mathcal{G}(s^*) = \{ (s, p, o) \in \mathcal{G} | s=s^* \vee o=s^*  \}$
\State $N_\mathcal{G}(o^*) = \{ (s, p, o) \in \mathcal{G} | s=o^* \vee o=o^*  \}$
\State $N_\mathcal{G}(s^*,o^*) = N_\mathcal{G}(s^*) \cup N_\mathcal{G}(o^*)$ \Comment{1-hop neighborhood of $s^*, o^*$}
\State $\mathcal{H} = \mathcal{H} \cup N_\mathcal{G}(s^*,o^*)$
\State $P_\mathcal{G}(p^*) = \{ (s, p, o) \in \mathcal{G} | p=p^*\}$ \Comment{Triples involving $p^*$}
\For{$i \gets 0$ \textbf{to} $n-1$}
\State Sample a triple $(\hat{s}, p^*, \hat{o}) \sim P_\mathcal{G}(p^*)$
\State Get the 1-hop neighborhood $N_\mathcal{G}(\hat{s},\hat{o})$
\State $\mathcal{H} = \mathcal{H} \cup N_\mathcal{G}(\hat{s},\hat{o})$
\EndFor
\end{algorithmic}
\end{algorithm}

\subsection{Knowledge Distillation} \label{method:kd}

After sampling the subgraph, we train a KGE model $g_S$ on that. We need to ensure that this model is a faithful surrogate to the black-box model whose predictions we are trying to explain, through some sort of constraint. 
For this reason, we propose the use of KD as a way to allow the original model to drive the learning process of the surrogate model. Thus, in KD terms, the black-box model plays the role of the teacher, while the surrogate model that of the student, with functions $g_T$ and $g_S$, respectively. The relational aspect of RKD makes it a natural fit for KGEs. Nevertheless, it is important to note that RKD is applied on individual samples. In mini-batch training, for instance, it will be applied on every possible combination of the samples that constitute the mini-batch. In KGEs, however, the relational aspect between the embeddings is inherent. Given that the training samples are in the form of triples $(s, p, o) \in \mathcal{G}$, the KGE loss is already affecting their embeddings $(\mathbf{s}, \mathbf{p}, \mathbf{o})$ in a relational manner. To accommodate the training of the KGE we adapt the RKD loss, so that it is applied only among the entities and the relation of a particular triple at a time, instead of randomly selected samples. We term this adaptation $\mathcal{L}_{RKD-KGE}$ and it takes the following form (Eq. \ref{eq:rkd_kge}):

\begin{align*}
\mathcal{L}_{RKD-KGE} = &\smashoperator{\sum_{(s,p,o) \in \mathcal{G}}}
l_\delta (\phi_A(\mathbf{s}_T,\mathbf{p}_T,\mathbf{o}_T), \phi_A(\mathbf{s}_S,\mathbf{p}_S,\mathbf{o}_S)) \\
&+l_\delta (\phi_A(\mathbf{p}_T,\mathbf{o}_T,\mathbf{s}_T), \phi_A(\mathbf{p}_S,\mathbf{o}_S,\mathbf{s}_S)) \\
&+l_\delta (\phi_A(\mathbf{o}_T,\mathbf{s}_T,\mathbf{p}_T), \phi_A(\mathbf{o}_S,\mathbf{s}_S,\mathbf{p}_S)) \numberthis
\label{eq:rkd_kge}
\end{align*}

The overall loss of the architecture can then be defined as in Eq. \ref{eq:loss}, with the RKD-KGE part acting as a regularization on the exact embeddings that the KGE part is affecting:

\begin{equation}
\mathcal{L} = \mathcal{L}_{KGE} + \lambda \mathcal{L}_{RKD-KGE}
\label{eq:loss}
\end{equation}

It is important to remind here that the teacher model is pre-trained and its weights are not updated. Instead the teacher representations are only utilized to aid the training of the student. An overview of the student's training procedure can be found in Fig. \ref{fig:arch}.

\subsection{Monte Carlo process} \label{method:mc}
The explanation produced by KGEx is a list of the triples included in the subgraph, ranked by their contribution to the prediction of the target triple. To generate this list, for each target triple that we want to explain and given a pre-trained KGE model (teacher), we train multiple KGE models (students) with the loss defined in Eq.~\ref{eq:loss}. Each of the students is trained on a subset $\mathcal{H}_{mc} \subset \mathcal{H}$ and assigns a \textit{rank} to the target triple. The contribution of each triple $t=(s,p,o) \sim \mathcal{H}$ is analogous to the average rank that was assigned to the target triple on the runs that $t$ was part of $\mathcal{H}_{mc}$.

\section{Experiments}\label{sec:eval}
We assess the faithfulness of the explanations returned by KGEx. Experiments show that KGEx surrogates are faithful to the original black-box models being explained.

\textbf{Datasets.}
We experiment with the two standard link prediction benchmark datasets, WN18RR~\cite{DBLP:conf/aaai/DettmersMS018} (a subset of Wordnet) and FB15K-237~\cite{toutanova2015representing} (a subset of Freebase). 
We operate with reduced test sets that include 100 triples only. This is to guarantee a reasonable execution time of our experiments, most of which require to retrain a model multiple times as part of the Monte Carlo step, for each triple that we want to explain.
We work with two separate test sets: first, to control for the black-box model predictive power, we define $\mathcal{T}_1$, that includes 100 test triples that have been ranked at the first place by all the black-box models used in our experiments (see `Evaluation Protocol' below).
Additionally, in some experiments we use another test set, $\mathcal{T}_{rand}$, which includes 100 randomly-selected triples, regardless of the assigned rank by the black-box KGE model.
Table~\ref{table:kgstats} shows the statistics of all the datasets used.

\begin{table}[t]
  \centering
  \footnotesize
  \begin{tabular}{l rr}
      \toprule          & FB15K-237 & WN18RR \\ 
      \midrule
      Training     &  272,115  &  86,835              \\ 
      Validation     &   17,535  &  3,034              \\ 
    $\mathcal{T}_1$: Test - Rank 1     &  100   &   100       \\
    $\mathcal{T}_{rand}$: Test - Random Rank   &  100   &   100       \\
      Entities         &  14,541   &  40,943    \\ 
      Relations          &  237   &  11     \\ 
      \bottomrule
  \end{tabular}
  \caption{Specifications of the datasets used in experiments}
  \label{table:kgstats}
\end{table}

\textbf{Evaluation protocol.}
We measure the faithfulness of explanations generated by KGEx in terms of predictive power discrepancy between black box predictions and predictions generated with KGEx surrogates.
We adopt the standard evaluation protocol described by \cite{bordes2013translating}. We predict whether each triple $t=(s,p,o) \in  \mathcal{T}$ is true, where $\mathcal{T}$ is either $\mathcal{T}_1$ or $\mathcal{T}_{rand}$. %
We cast the problem as a learning-to-rank task: for each $t=(s,p,o) \in \mathcal{T}$, we generate synthetic negatives $t^- \in \mathcal{N}_t$ by corrupting one side of the triple at a time (i.e. either the subject or the object). In the standard evaluation protocol, synthetic negatives $\mathcal{N}_t$ are generated from all entities in $\mathcal{E}$. In our experiments, to guarantee a fair comparison between the black-box and the surrogate model, we limit to synthetic negatives created from entities included in the corresponding sampled subgraph $\mathcal{H}$.
We predict a score for each $t$ and all its negatives $t^- \in \mathcal{N}_t$. We then rank the only positive $t$ against all the negatives $\mathcal{N}_t$. 
We report mean rank (MR), mean reciprocal rank (MRR), and Hits at $n$ (where $n=1, 10$) by filtering out spurious ground truth positives from the list of generated corruptions (i.e. ``filtered'' metrics).

\textbf{Implementation Details and Baselines.}
The KGEx explanation subsystem and the black-box KGE models are implemented using TensorFlow 2.5.2 and Python 3.8\footnote{We will release code and experiments upon acceptance.}.
KGE hyperparameter ranges and best combinations are reported in Appendix~\ref{app:hyperparam}.
Regarding the KGEx specific hyperparameters, we use Predicate Neighborhood (PN) sampling with 5 neighbors for FB15K-237 and 3 neighbors for WN18RR (see section \ref{exp:subgraph}), KD coefficient $\lambda=3$ (see section \ref{exp:kd}). Student models have embedding dimensionality $k=50$, synthetic negatives ratio $\eta=2$ and are trained using the Adam optimizer and a multiclass-NLL baseline loss with learning rate=0.1 for 200 epochs. 
Depending on the size of the subgraph, an explanation might require from 50 to 200 MC runs.
All experiments were run under Ubuntu 16.04 on an Intel Xeon E5-2630, 32 GB, equipped with a Titan XP 12GB.

\subsection{Faithfulness: KGE Architectures}\label{exp:fidelity}
The first experiment tests how faithful the KGEx surrogates are to the pre-trained black-box models in terms of predictive performance. We conduct this experiment by leveraging both the $\mathcal{T}_1$ test set (which is produced from the ranks of each black-box that we are explaining) and the $\mathcal{T}_{rand}$ test set. 

The results for $\mathcal{T}_1$ are shown on Table \ref{table:exp1.1a-2.1a}. Naturally, all black-box models have perfect metrics, by definition. By inspecting the performance of the surrogate models, we can see that they retain a respectable level in MRR of around 0.5 (depending on the black-box) for FB15K-237. It is quite interesting to note that the TransE surrogate is the one which manages to achieve the minimum drop in performance from its black-box, across all reported metrics. Turning to WN18RR, we can see even stronger results from the surrogates, and especially from DistMult and ComplEx. DistMult in particular replicates its black-box's almost perfect performance.

The same experiment is conducted on $\mathcal{T}_{rand}$ test set and the results are reported on Table \ref{table:exp1.1b-2.1b}. 
TransE again gets the best results for FB15K-237 with 53\% drop in MRR from its black-box, but given the added difficulty of the task, DistMult and ComplEx still manage to get a comparable Hits@10. For WN18RR, similar to $\mathcal{T}_{1}$, DistMult and ComplEx retain quite high scores across all the metrics. While TransE is a bit lower compared to the other architectures, we can see that compared to its own black-box, it actually manages to stay on a very competitive level in terms of MRR and Hits@10. We have to mention here that the MRR of the TransE black-box in this case is 35-40\% lower than DistMult or ComplEx, which is confirmed by similar numbers reported in recent literature. The reason for that is most likely related to the small amount of relations in WN18RR, which might not be properly captured by TransE's architecture. As a result, this affects the TransE surrogate both in $\mathcal{T}_{1}$ and $\mathcal{T}_{rand}$ and therefore is \textit{not} an issue of the KGEx component but rather the black-box's.

\begin{table*}
  \centering
    \small

  \begin{tabular}{ll @{\extracolsep{8pt}} cccc  cccc}
      & & \multicolumn{4}{c}{\textbf{FB15K-237 - Rank 1 ($\mathcal{T}_1$)}}   & \multicolumn{4}{c}{\textbf{WN18RR - Rank 1 ($\mathcal{T}_1$)}}       \\
      \cline{3-6} \cline{7-10}

      & & & & \multicolumn{2}{c}{Hits@}  &    & & \multicolumn{2}{c}{Hits@}    \\
      \cline{5-6} \cline{9-10}

      & & MR & MRR & 1  & 10 & MR & MRR & 1  & 10  \\

       \midrule

        \multicolumn{2}{l}{TransE} 
        & 1      & 1.0          & 1.0       & 1.0
        & 1     & 1.0          & 1.0       & 1.0  \\

        \multicolumn{2}{l}{DistMult}  
        & 1      & 1.0         & 1.0       & 1.0
        & 1      & 1.0      & 1.0       & 1.0     \\

        \multicolumn{2}{l}{ComplEx} 
        & 1      & 1.0          & 1.0      & 1.0
        & 1      & 1.0      & 1.0       & 1.0     \\
        
        \midrule

        \multirow{3}{*}{\textbf{KGEx}}
        & TransE
        & \textbf{59}      & \textbf{.55}       & .44      & \textbf{.75}        
        & 41      & .24      & .12     & .52     \\
        
        & DistMult
        & 72      & .54      & \textbf{.46}      & .69         
        & \textbf{5}      & \textbf{.99}      & \textbf{.98}      & \textbf{.98}    \\
        
        & ComplEx
        &   130    & .50      & .42       & .65     
        & 8      & .87      & .84       & .92    \\

  \end{tabular}
  \caption*{}

\caption{Faithfulness on $\mathcal{T}_1$ (triples that were ranked 1 by the black-box): comparison between KGE architectures. Worst possible rank is 2712 for FB15K-237 and 370 for WN18RR. Filtered metrics. Best results in bold.}
\label{table:exp1.1a-2.1a}
\end{table*}

\begin{table*}
  \centering
    \small

  \begin{tabular}{ll @{\extracolsep{8pt}} cccc  cccc}
      & & \multicolumn{4}{c}{\textbf{FB15K-237 - Random Rank ($\mathcal{T}_{rand}$)}}   & \multicolumn{4}{c}{\textbf{WN18RR - Random Rank ($\mathcal{T}_{rand}$)}}       \\
      \cline{3-6} \cline{7-10}

      & & & & \multicolumn{2}{c}{Hits@}  &    & & \multicolumn{2}{c}{Hits@}    \\
      \cline{5-6} \cline{9-10}

      & & MR & MRR & 1  & 10 & MR & MRR & 1  & 10  \\

       \midrule

        \multicolumn{2}{l}{TransE} 
        & 34      & .36          & .24       & .61
        & \textbf{6}     &   .36        & .10     & \textbf{.85}  \\

        \multicolumn{2}{l}{DistMult}  
        & 36     & .37        & .24       & \textbf{.64}
        & 21     & .55      & .47       & .70     \\

        \multicolumn{2}{l}{ComplEx} 
        & \textbf{27}      & \textbf{.41}          & \textbf{.30}      & .62
        & 24      & \textbf{.61}      & \textbf{.49}       & \textbf{.78}    \\
        
        \midrule

        \multirow{3}{*}{\textbf{KGEx}}
        & TransE
        & \textbf{118}     &   \textbf{.17}        & \textbf{.10}       & \textbf{.27}  
        & 39      & .19      & .00     & .55     \\
        
        & DistMult
        & 284      & .13     & .06       & .25    
        & \textbf{20}      & \textbf{.43}      & \textbf{.38}      & \textbf{.57}    \\
        
        & ComplEx
        & 282      & .12      & .05       & .26   
        & 36      & .39     & .33      & .47    \\

  \end{tabular}
  \caption*{}

\caption{Faithfulness on $\mathcal{T}_{rand}$ (randomly selected triples regardless of the black-box's assigned rank): comparison between KGE architectures.  Worst possible rank is 2712 for FB15K-237 and 370 for WN18RR. Filtered metrics. Best results in bold.}
\label{table:exp1.1b-2.1b}
\end{table*}

\begin{table}
  \centering
  \footnotesize

  \begin{tabular}{lccccc}
      & \multicolumn{5}{c}{\textbf{FB15K-237 - Rank 1 ($\mathcal{T}_1$)}}     \\
      \cline{2-6} 

      Subgraph Sampling & Subgraph& && \multicolumn{2}{c}{Hits@}     \\
      \cline{5-6}

        & Avg size & MR & MRR & 1  & 10   \\

       \midrule
        
        3 Predicate Neighbors
        & 750 & 113 & .40 & .32 & .53 \\
        
        5 Predicate Neighbors
        & 1783 & 130 & \textbf{.50} & \textbf{.42} & .65 \\
        
        10 Predicate Neighbors
        & 2742 & 171 & .47 & .38 & \textbf{.66} \\
        
        \midrule
        
        10 Random Walk Steps
        & 521 & \textbf{99} & .23 & .09 & .43 \\
        
        50 Random Walk Steps
        & 555 & 124 & .21 & .11 & .36    \\
        
        100 Random Walk Steps
        & 606 & 150 & .20 & .10 & .37 
    
  \end{tabular}

  \vspace{.5cm}

  \begin{tabular}{lccccc}
      & \multicolumn{5}{c}{\textbf{WN18RR - Rank 1 ($\mathcal{T}_1$)}}     \\
      \cline{2-6} 

      Subgraph Sampling & Subgraph& && \multicolumn{2}{c}{Hits@}     \\
      \cline{5-6}

        & Avg size & MR & MRR & 1  & 10   \\

       \midrule
        
        3 Predicate Neighbors
        & 61 & 8 & \textbf{.87} & \textbf{.84} & \textbf{.92} \\
        
        5 Predicate Neighbors
        & 74 & \textbf{6} & .84 & .81 & .90\\
        
        10 Predicate Neighbors
        & 132 & 8 & .81 & .77 & .89 \\
        
        \midrule
        
        100 Random Walk Steps
        & 86 & \textbf{6} & .68 & .57 & .87 \\
        
        200 Random Walk Steps
        & 153 &8 & .63 & .49 & .84    \\
        
        500 Random Walk Steps
        & 351 &23 & .55 & .41 & .79 

  \end{tabular}
  \caption*{}

\caption{Subgraph Sampling approaches. KGE architecture used is ComplEx. Filtered metrics. Best results in bold.}
\label{table:exp1.2-2.2}
\end{table}

\begin{table*}
  \centering
    \small

  \begin{tabular}{ll @{\extracolsep{8pt}} cccc  cccc}
      & & \multicolumn{4}{c}{\textbf{FB15K-237 - Rank 1 ($\mathcal{T}_1$)}}   & \multicolumn{4}{c}{\textbf{WN18RR - Rank 1 ($\mathcal{T}_1$)}}       \\
      \cline{3-6} \cline{7-10}

      &Knowledge & & & \multicolumn{2}{c}{Hits@}  &    & & \multicolumn{2}{c}{Hits@}    \\
      \cline{5-6} \cline{9-10}

      & Distillation & MR & MRR & 1  & 10 & MR & MRR & 1  & 10  \\

       \midrule

        TransE
        & \multirow{3}{*}{No}
        & \textbf{37}      & .54      & \textbf{.46}       & .69        
        & 27      & .38      & .27     & .60     \\

        Distmult
        &
        & 157      & .40         &.31      & .55
        & 7     & .98      & \textbf{.98}       & \textbf{.98}     \\

        ComplEx
        &
        & 189      & .37          & .28      & .55
        & 8      & .81      & .76      & .90    \\
        
        \midrule

        TransE
        & \multirow{3}{*}{Yes}
        & 59    &   \textbf{.55}        & .44       & \textbf{.75}
        & 41      & .24      & .12     & .52     \\
        DistMult
        & 
        & 72      & .54      & \textbf{.46}      & .69         
        & \textbf{5}      & \textbf{.99}      & \textbf{.98}      & \textbf{.98}    \\
        ComplEx
        & 
        &   130    & .50      & .42       & .65     
        & 8      & .87      & .84       & .92    \\

  \end{tabular}
  \caption*{}

\caption{Impact of Knowledge Distillation on KGEx: comparison between KGE architectures. Filtered metrics. Best results in bold.}
\label{table:exp1.3-2.3}
\end{table*}

\begin{figure}
\centering
\includegraphics[scale=0.5]{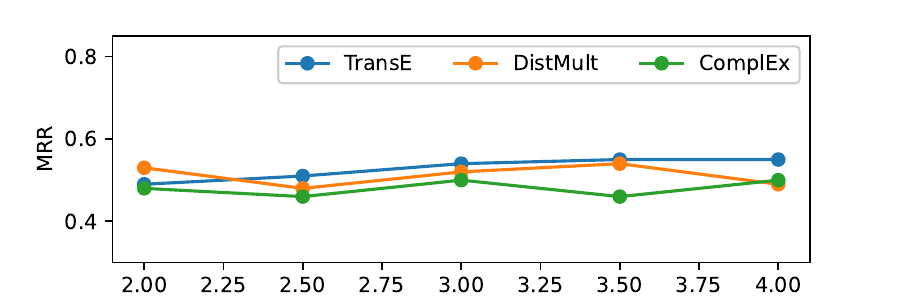}
\caption{Impact of Knowledge Distillation loss coefficient $\lambda$ on model performance for TransE and ComplEx on FB15K-237.}
\label{fig:kd_coef}
\end{figure}

\subsection{Impact of Subgraph Sampling}\label{exp:subgraph}
We evaluate predicate neighborhood (PN) and random walk (RW) sampling. Both methods contain the 1-hop neighborhood of the subject and the object of the target triple. For PN sampling, we conduct experiments with 3, 5 and 10 predicate neighbors for both datasets. For RW sampling, we use 10, 50 and 100 steps for FB15K-237 and 100, 200 and 500 steps for WN18RR. Table~\ref{table:exp1.2-2.2} shows complete results. 

Across both datasets PN sampling yields much better results than RW sampling. An interesting finding, is that regardless of method increasing the subgraph size either with more neighbors for PN sampling or with more steps in RW sampling, performance is actually decreasing. This shows that we do not need a very large subgraph to calculate effective embeddings for the target triple, and that including facts likely to be not relevant in the subgraph hurts performance.

\subsection{Knowledge Distillation Effect}\label{exp:kd}
We assess the effect of knowledge distillation on the KGEx pipeline. As defined in Eq.~\ref{eq:loss}, the KD component acts as regularization in the training process of the student. Its contribution is regulated by the KD coefficient $\lambda$ and, as we see in Fig.~\ref{fig:kd_coef}, performance across models remains fairly stable across different $\lambda$. We chose $\lambda=3$ across experiments, as it consistently gives marginally better results across all models. 

Finally, we look at the enhancement in performance that KD offers in Table~\ref{table:exp1.3-2.3}. It is evident that when KD is used, it improves results across all backbone models, compared to standalone students (i.e. models trained on the subgraph without KD). In detail, ComplEx and DistMult outperform their standalone counterparts across all metrics for both datasets with increase in MRR in the 6-13 \% region. The outlier in our observations comes from TransE on WN18RR, which is the only case where the standalone model outperforms the KD-based one (MRR=0.38 against MRR=0.24 on $\mathcal{T}_1$). 
This is \textit{not} a limitation of KGEx though, but rather something that highlights it is indeed working as intended. Given the characteristics of WN18RR (small number of relations), it looks like a smaller subgraph actually favors KGEs, as the standalone model, even though of smaller capacity, outperforms the model trained on the full KG. However, the KD student remains bounded by the knowledge that was distilled by its teacher and therefore remains faithful to the teacher, which is what is desirable. On the other hand, the standalone model leverages to a great extent the favorable topology of the subgraph, but because there is no connection to the teacher through the loss function, it fails completely to capture any of the teacher's representation abilities, which is what we would expect.

\subsection{Example Explanations}\label{exp:examples}
We also provide some example explanations in Table \ref{table:examples}. These were generated by KGEx for target triples from FB15K-237. Something that can be observed is KGEx's ability to include triples in the explanation that are beyond the 1-hop neighborhood of the subject and the object of the target triple. Such an example is the explanation triple \textit{(Julianne Moore, film, Far from heaven)}, which explains the target triple \textit{(Walk of fame, inductee, Meryl Streep)} because Meryl Streep and Julianne Moore have collaborated in movies. When explaining the target triple \textit{(Priyanka Chopra, ethnicity, Punjabis)}, we see that the explanation contains relevant information and not facts about the subject's professional life as in the previous example. Finally, in other cases, when the context can be really specific, such as the symptoms example, we see that all the explanation triples are sourced based on the predicate. Even in that case though, although jaundice can be a symptom of various diseases, the ones chosen in the explanation (i.e. pancreatic cancer and malaria) are both correlated with hepatitis.

\setlength{\tabcolsep}{1pt}
\begin{table*}
  \centering
    \small
\begin{tabular}{lccc@{\extracolsep{5pt}}lccc}
&\textbf{Walk of fame} & \textbf{inductee} & \textbf{Meryl Streep} & &\textbf{Ryanair} &\textbf{headquarters} &\textbf{Dublin} \\
\cline{1-4} \cline{5-8}
1&Meryl Streep& award& Tony award&1&Ryanair&currency&Euro \\
2&Julianne Moore&film&Far from heaven&2&Ryanair&phone service&Ireland\\
3&84th Academy awards&winner&Meryl Streep&3&Dublin&transportation&Air travel\\
\cline{1-4} \cline{5-8} \\
&&&&&&& \\
&\textbf{Jaundice} & \textbf{symptom of} & \textbf{Hepatitis} &&\textbf{Priyanka Chopra} &\textbf{ethnicity} &\textbf{Punjabis} \\
\cline{1-4} \cline{5-8}
1&Jaundice & symptom of & Pancreatic cancer & 1 & Punjabis & location & Pakistan \\
2&Abdominal pain & symptom of & Hepatitis& 2 & Priyanka Chopra & lived & Jharkhand \\
3&Jaundice & symptom of & Malaria & 3 & Juhi Chawla & ethnicity & Punjabis\\
\end{tabular}

\caption{Example explanations for FB15K-237 by KGEx. Each target triple is noted in bold. All triples are predicted as factually correct with a prediction score of 0.99. The top three explanation triples below each target triple are listed in order of importance. The black-box model is ComplEx with embedding dimentionality $k=350$. The student models' embedding dimentionality is $k=50$. The subgraph sampling approach is random walk with 10 steps. The number of Monte Carlo runs is 100 and in each run the subgraph of each target triple is partitioned in 10 subsets.}
\label{table:examples}
\end{table*}

\section{Conclusion}\label{sec:conclusion}
KGEx generates post-hoc, local explanations in the form of a ranked list of triples. We show that the interplay of graph sampling and knowledge distillation reduces the explanation search space while guaranteeing faithfulness to the black-box KGE model being explained. Deploying a Monte Carlo process to rank the explanation triples based on their influence on the prediction prioritizes important and relevant facts.

Moving forward, we will leverage the modular nature of the framework and propose replacement modules, such as a search-based approach instead of the MC process, to reduce computational burden.
Future work will also focus on user studies to gauge the perceived quality of KGEx explanations, by measuring how much they assist human experts on the receiving side of KGE predictions.

\bibliographystyle{named}
\bibliography{references}

\appendix
\clearpage

\section*{KGEx: Explaining Knowledge Graph Embeddings via Subgraph Sampling and Knowledge Distillation: Supplementary material}

\section{Subgraph sampling with random walk}\label{app:random_walk_algorithm}

\begin{algorithm}
\caption{Subgraph sampling w/ Random Walk}\label{alg:random_walk_subgraph}
\begin{algorithmic}[1]
\State \textbf{Input:} {target triple $(s^*,p^*,o^*)$, number of random walk steps $n$}
\State \textbf{Output:} {Subgraph $\mathcal{H}$}
\State $\mathcal{H} \gets \emptyset$
\State $N_\mathcal{G}(s^*) = \{ (s, p, o) \in \mathcal{G} | s=s^* \vee o=s^*  \}$
\State $N_\mathcal{G}(o^*) = \{ (s, p, o) \in \mathcal{G} | s=o^* \vee o=o^*  \}$
\State $N_\mathcal{G}(s^*,o^*) = N_\mathcal{G}(s^*) \cup N_\mathcal{G}(o^*)$ \Comment{1-hop neighborhood of $s^*, o^*$}
\State $\mathcal{H} = \mathcal{H} \cup N_\mathcal{G}(s^*,o^*)$
\State $(s_o, p_o, o_o) = (s^*, p^*, o^*)$ \Comment{Initialize random walk origin}
\For{$i \gets 0$ \textbf{to} $n-1$}
\State Sample a triple $(\hat{s}, \hat{p}, \hat{o}) \sim N_\mathcal{G}(s_o,o_o)$
\State $\mathcal{H} = \mathcal{H} \cup \{(\hat{s}, \hat{p}, \hat{o})\})$
\State $(s_o, p_o, o_o) = (\hat{s}, \hat{p}, \hat{o}) $ \Comment{Update origin}
\EndFor
\end{algorithmic}
\end{algorithm}

\section{Hyperparameter search}\label{app:hyperparam}
We experiment with three popular KGE architectures: TransE, DistMult, ComplEx. 
For each of them we replicated SOTA results by carrying out extensive grid search, over the following ranges of hyperparameter values: embedding dimensionality $k=[200-500]$, with a step of 50; baseline losses=\{negative log-likelihood, multiclass-NLL, self-adversarial\}; synthetic negatives ratio $\eta=\{20, 30, 40, 50\}$; learning rate$=\{\num{1e-4}, \num{5e-5}, \num{1e-5}\}$; epochs$=[500-2000]$, step of 500; L2 regularizer, with weight $\gamma=\{\num{1e-3}, \num{1e-4}, \num{1e-5}\}$. The best loss for all models was the multiclass-NLL and the best regularization weight $\gamma=\num{1e-4}$. The best combinations for the rest of the hyperparameters are shown on Table \ref{table:hparams}

\begin{table*}[h!]
  \centering
    \small

  \begin{tabular}{l @{\extracolsep{8pt}} cccc  cccc}
       & \multicolumn{4}{c}{\textbf{FB15K-237}}   & \multicolumn{4}{c}{\textbf{WN18RR}}       \\
      \cline{2-5}  \cline{6-9}

      & k & $\eta$ & lr & epochs & k & $\eta$  & lr & epochs \\

       \cline{1-5} \cline{6-9}

        TransE
        
        & 400      & 30   & 1e-4     
        & 1000          &  350   &   30   & 1e-4 & 2000\\

        Distmult
        
        & 300      & 50     & 5e-5
        & 1000       &   350     &  30  & 1e-4 & 2000  \\

        ComplEx
        
        & 350      & 30     & 5e-5
        & 1000      &   200   & 20   & 5e-5 & 2000 \\

  \end{tabular}
  \caption*{}

\caption{Best hyperparameter combinations for baseline black-box models. }
\label{table:hparams}
\end{table*}

\end{document}